\documentclass[11pt,a4paper]{article}
\usepackage{authblk}
\usepackage[hyperref]{emnlp2020}
\usepackage{enumitem,microtype,times,latexsym,booktabs,mdwlist,natbib,xcolor,hyperref,float,tipa,subcaption,linguex,soul,graphicx}

\aclfinalcopy 
\def\aclpaperid{3255} 

\title{
What's so special about BERT's layers? \\ A closer look at the NLP pipeline in monolingual and multilingual models
}

\author[]{\textbf{Wietse de Vries}}
\author[]{\textbf{Andreas van Cranenburgh}}
\author[]{\textbf{Malvina Nissim}}

\affil{CLCG, University of Groningen, The Netherlands
    \vspace{5px} \\
    \texttt{\{wietse.de.vries,a.w.van.cranenburgh,m.nissim\}@rug.nl} 
} 

\date{}

\begin{document}
\maketitle

\begin{abstract}
Peeking into the inner workings of BERT has shown that its layers resemble
the classical NLP pipeline,
with progressively more complex tasks being concentrated in later layers.
To investigate to what extent these results also hold for a language other than English, we probe a Dutch BERT-based model and the multilingual BERT model for Dutch NLP tasks.
In addition, through a deeper analysis of part-of-speech tagging, we show that also within a given task, information is
spread over different parts of the network and the pipeline might not be as neat as it seems. Each layer has different specialisations, so that it may be more useful to combine information from different layers, instead of selecting a single one based on the best overall performance.
\end{abstract}

\section{Introduction and Background}
Natural Language Processing is now dominated by transformer-based models \citep{vaswani2017attention}, like BERT \citep{devlin2019bert}, a model trained on predicting masked tokens and relations between sentences. BERT's impact is so strong that we already talk about `BERTology' \citep{bertology}. 

In addition to using BERT in NLP tasks and end applications, research has also been done \textit{on} BERT, especially to reveal what linguistic information is available in different parts of the model. This is done, e.g., investigating 
what BERT's attention heads might be attending to \citep{clark-etal-2019-bert}, or looking at its internal vector representations using so-called probing (or diagnostic) classifiers \citep{tenney2019bert}.
It has been noted that BERT  progressively acquires linguistic information roughly in the same the order of the classic language processing pipeline \citep{tenney2019you,tenney2019bert}: surface features are expressed in lower layers, syntactic features more in middle layers and semantic ones in higher layers \citep{jawahar-etal-2019-bert}. So, for example, information on part-of-speech appears to be acquired earlier than on coreference.

Most work dedicated to understanding the inner workings of BERT has focused on English, though non-English BERT models do exist, in two forms.
One is a multilingual model \citep[mBERT]{devlin2019bert}, which is trained on Wikipedia dumps of 104 different languages. The other one is a series of monolingual BERTs \citep[among others]{polignano2019alberto, le2019flaubert, virtanen2019multilingual, camembert, bertje}. As expected, also the non-English monolingual BERT models achieve state-of-the-art results on a variety of NLP tasks, and mostly outperform the multilingual model on common NLP tasks \citep{nozza2020mask}. Nevertheless, mBERT performs surprisingly well on zero-shot POS tagging and Named Entity Recognition (NER), as well as on cross-lingual model transfer \citep{pires-etal-2019-multilingual}.

If these results imply that the inner workings of other monolingual BERTs and of mBERT are the same as BERT's is not yet known. Also not known is how \textit{homogeneous}  layer specialisation is: through general performance of, e.g., POS tagging, we see a peak at a given layer, but we do not know how specialisation actually evolves across the whole model. 
This work investigates such issues. 

\paragraph{Contributions}
Using probing classifiers for four tasks on six datasets for a monolingual Dutch model and for mBERT, we observe that (i) these models roughly exhibit the same classic pipeline observed for the original BERT, suggesting this is a general feature of BERT-based models; (ii) the most informative mBERT layers are consistently earlier layers than in monolingual models, indicating an inherent task-independent difference between the two models. Through a deeper analysis of POS tagging, we also show that (iii) the picture of a neatly ordered NLP pipeline is not completely correct, since information appears to be more spread across layers than suggested by the performance peak at a given layer.

The full source code is publicly available on Github\footnote{\url{https://github.com/wietsedv/bertje/tree/master/probing}}.

\section{Approach}

We run two kinds of analyses. 

The first is aimed at a rather high level comparison of the performance of a monolingual (Dutch) BERT model (BERTje, \citealt{bertje}) and multilingual BERT (mBERT) on a variety of tasks at different levels of linguistic complexity (POS tagging, dependency parsing, named entity recognition, and coreference resolution; see Section~\ref{sec:task-data}), with attention to what happens at different layers. 

The second is an in-depth analysis of the performance of BERTje and mBERT on part-of-speech tagging. The reason behind this is that looking at global performance over a given task does not provide enough information on what is actually learned by different layers of the model \textit{within} that task. POS tagging lends itself well for this type of layerwise evaluation.
First, because it is a low level task for which relatively little real-world knowledge is required.
Second, because analysis of single tags is straightforward since it is done at a token level.
Third, because POS tagging contains both easy and difficult cases that depend on surrounding context. Some words are more ambiguous than others, and some classes are open whereas others are closed.
Token ambiguity may for instance be an important factor for differences between a monolingual and a multilingual model since the latter has to deal with more homographs, due to the co-presence of multiple languages.

Section~\ref{sec:analysis} describes how these analyses can be performed in practice using the probes.

\subsection{Experimental setup}
Our method for measuring task performance at different layers is based on the edge probing approach of \citet{tenney2019bert, tenney2019you}.
Edge probing is a method to evaluate how well linguistic information can be extracted from a pre-trained encoder.
Separate trained classifiers on the outputs of Transformer layers in BERT can reveal which layers contain most information for a particular  task. 

The inputs of the probing classifiers are embeddings extracted from the lexical layer (layer 0) and each Transformer layer (layers 1 up to 12) from either the pre-trained BERTje or mBERT model.
Embeddings of token spans are extracted from these full sentence or document embeddings and those spans are used as probe model inputs.
The probing classifiers are trained to predict task labels based on span representations using an LSTM layer for tokens that require multiple WordPieces.\footnote{See \citet{tenney2019bert} for technical details on the classifier architecture. Our hyper-parameters are in Appendix~\ref{appendix:hyperparams}.}

For each model, layer and task we train two probes: a single layer based probe and a scalar mixing probe.
The single layer probe uses a single pre-trained Transformer layer output as its input, whereas the scalar mixing probes use a weighted sum of the target layer and preceding layers.

\subsection{Tasks and Data}
\label{sec:task-data}

We train the probing classifiers on six datasets with four different tasks, chosen to represent linguistic layers of abstraction.\footnote{Details on size, splits, and processing are in Appendix~\ref{appendix:data}.}
For POS tagging and dependency parsing, the LassySmall and Alpino datasets from Universal Dependencies (UD) v2.5 \citep{ud25} are used with provided splits.
For Named Entity Recognition, we use the Dutch portion of the CoNLL-2002 NER dataset \citep{tjong2002conll} with the provided splits.
Finally, we use the coreference annotations of the SoNaR-1 corpus \citep{sonar1} for coreference, with document level training (80\%), validation (10\%) and testing (10\%) splits.

\begin{figure*}[!t]
  \begin{subfigure}[b]{0.5\textwidth}
    \includegraphics[width=\textwidth]{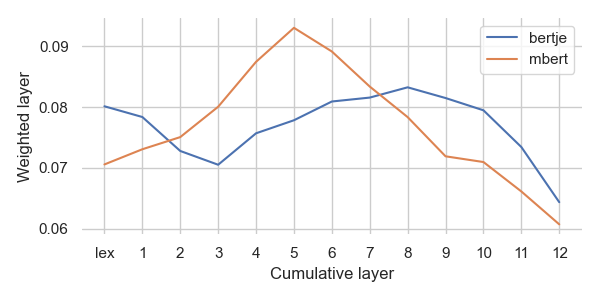}
    \caption{UDLassy POS}
    \label{fig:weights:udlassy-pos}
  \end{subfigure}
  \begin{subfigure}[b]{0.5\textwidth}
    \includegraphics[width=\textwidth]{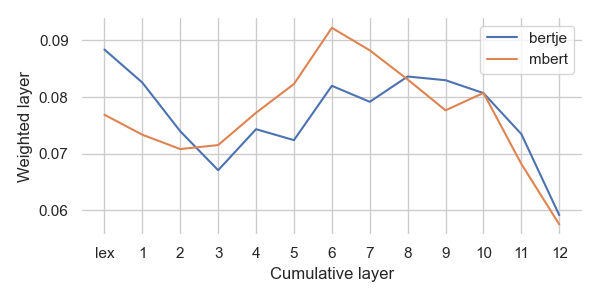}
    \caption{UDAlpino POS}
    \label{fig:weights:udalpino-pos}
  \end{subfigure}
  \begin{subfigure}[b]{0.5\textwidth}
    \includegraphics[width=\textwidth]{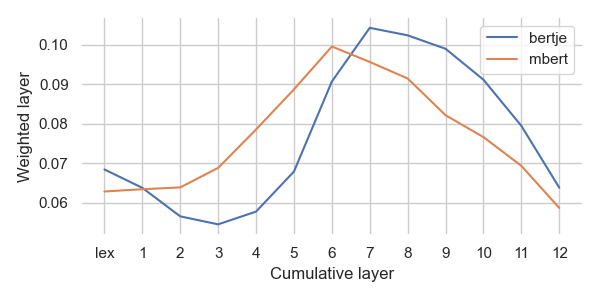}
    \caption{UDLassy DEP}
    \label{fig:weights:udlassy-dep}
  \end{subfigure}
  \begin{subfigure}[b]{0.5\textwidth}
    \includegraphics[width=\textwidth]{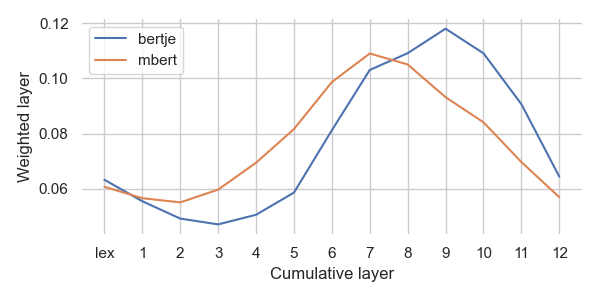}
    \caption{UDAlpino DEP}
    \label{fig:weights:udalpino-dep}
  \end{subfigure}
  \begin{subfigure}[b]{0.5\textwidth}
    \includegraphics[width=\textwidth]{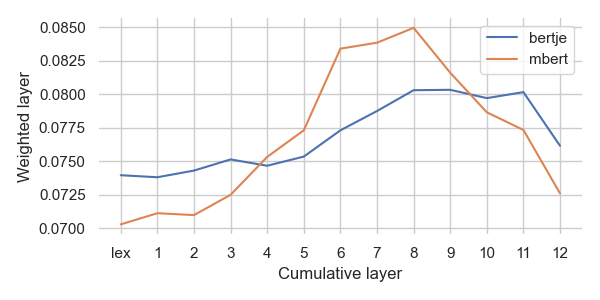}
    \caption{CoNLL-2002 NER}
    \label{fig:weights:conll2002-ner}
  \end{subfigure}
  \begin{subfigure}[b]{0.5\textwidth}
    \includegraphics[width=\textwidth]{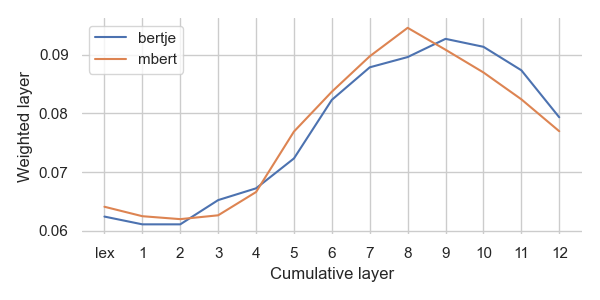}
    \caption{SoNaR Coref}
    \label{fig:weights:sonar-coref}
  \end{subfigure}
  \caption{Scalar mixing weights for each pre-trained model and each task.
  Highlights:
  The sorted weights form clean curves;
  BERTje makes more use of lexical embeddings;
  Weights  decrease  at  final  layers;
  mBERT peaks earlier than BERTje;
  POS and DEP results are consistent across datasets.
  }
  \label{fig:weights}
\end{figure*}

\subsection{Analysis}
\label{sec:analysis}
We perform a series of analyses aimed at creating a picture of what happens inside of BERTje and mBERT.
Initial overall analyses of the tasks are done with the scalar mixing probes as well as the single layer probes for each of the six tasks.

First, weights that the scalar mixing probes give to each pre-trained model layer are compared (Section~\ref{sec:weights}).
Layers that get larger scalar mixing weights may be considered to be more informative than lower weight layers for a particular task \citep{tenney2019bert}.
It does not have to be the case that the most informative layers are at the same position in the model since an interaction between layers in different positions may be even more informative.
Therefore, we compare layer weights between tasks and pre-trained models.
The two different data sources for POS tagging and dependency parsing will give an indication about stability of these weight distributions across datasets and within tasks.
These weights are solely based on training data, so they may not represent the exact layer importance for unseen data.

Second, we compare overall prediction scores of the probes on unseen test data for each task (Section~\ref{sec:predictions}).
Through this, we can observe at what stage models peak for what task, and where monolingual and multilingual models might differ.
The accuracy deltas between layers for scalar mixing probes will give an indication about which layers add information that was not present in all previous layers combined.
For these probes, deltas should be positive if information is added and zero if a layer is uninformative.

Third, we take a closer look at POS tagging (Section~\ref{sec:pos}). 
The previous analyses reveal information about the amount of task-relevant information that is present in each layer, but POS tagging can require different kinds of abstraction for different labels, so that POS performance might be non-homogeneous across layers.
Specifically, we (i) compare layerwise performance for each tag and the groups of open and closed class POS tags; (ii) 
investigate whether information is lost, learned or relearned within the  model by combining probe predictions for each individual token; and (iii)
check the most frequent confusions between tags to better understand the causes of errors.

\section{Analysis over all tasks}

First, the weights of the scalar mixing models are compared in order to see which layer combinations are most informative. 
These weights are tuned solely on the training data so they give no indication about layer importance for unseen data.
Second, we compare overall prediction scores of the probes on unseen test data for each of the tasks.

\subsection{Layer weights}\label{sec:weights}

Figure \ref{fig:weights} shows the scalar mixing weights of the full scalar mixing probes. We highlight a few important patterns that are consistent between tasks, and suggest possible explanations for what we observe, in particular regarding the differences between BERTje and mBERT.

\paragraph{The sorted weights form clean curves.}
The probing classifier is ignorant about ordering of layers when the weights are tuned.
Nevertheless the sorted weights mostly show clean curves.
The clean curves indicate that embedding of useful information for these tasks is gradually added and removed by the transformer models.
This also confirms that our probing model is actually sensitive to these gradual changes in the embeddings.

\begin{figure*}[!t]
\centering
  \begin{subfigure}[b]{0.45\textwidth}
    \includegraphics[width=\textwidth]{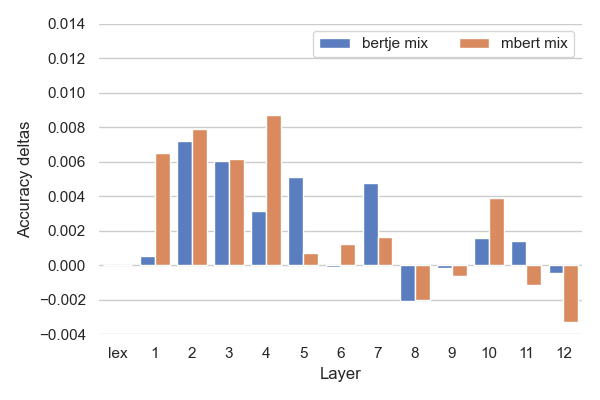}
    \caption{UDLassy POS}
  \end{subfigure}
  \begin{subfigure}[b]{0.45\textwidth}
    \includegraphics[width=\textwidth]{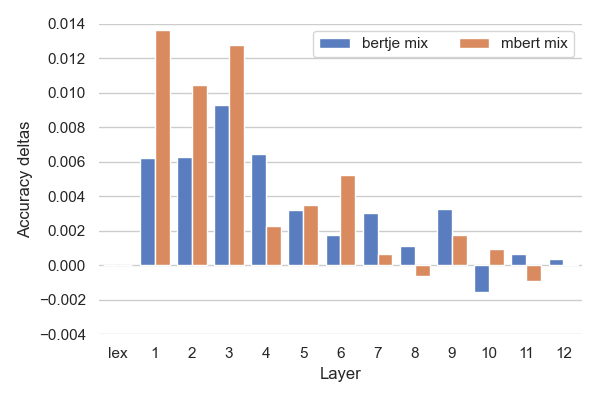}
    \caption{UDAlpino POS}
  \end{subfigure}
  \begin{subfigure}[b]{0.45\textwidth}
    \includegraphics[width=\textwidth]{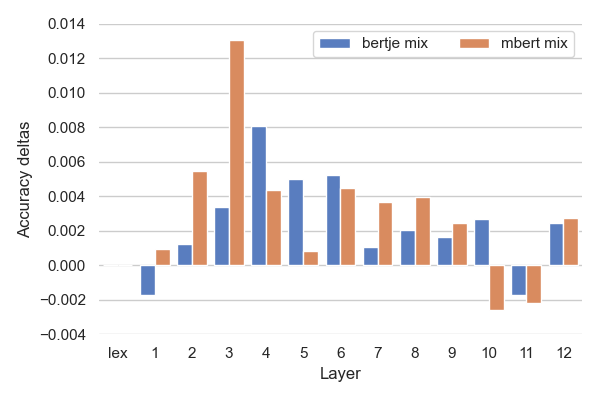}
    \caption{UDLassy DEP}
  \end{subfigure}
  \begin{subfigure}[b]{0.45\textwidth}
    \includegraphics[width=\textwidth]{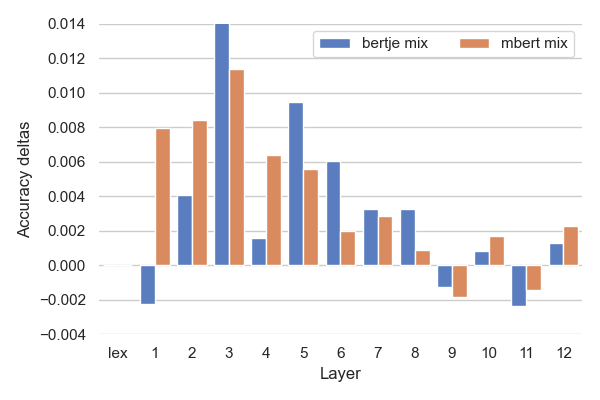}
    \caption{UDAlpino DEP}
  \end{subfigure}
  \begin{subfigure}[b]{0.45\textwidth}
    \includegraphics[width=\textwidth]{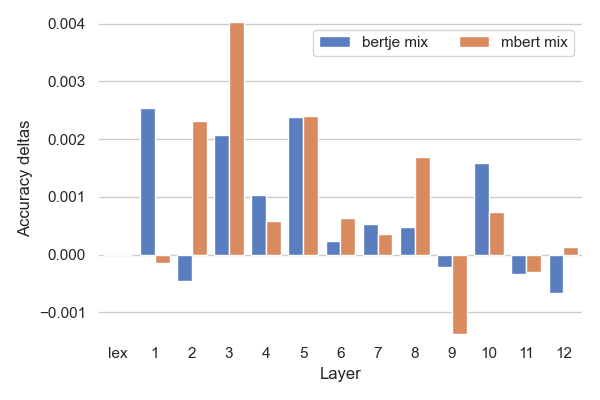}
    \caption{CoNLL-2002 NER}
  \end{subfigure}
  \begin{subfigure}[b]{0.45\textwidth}
    \includegraphics[width=\textwidth]{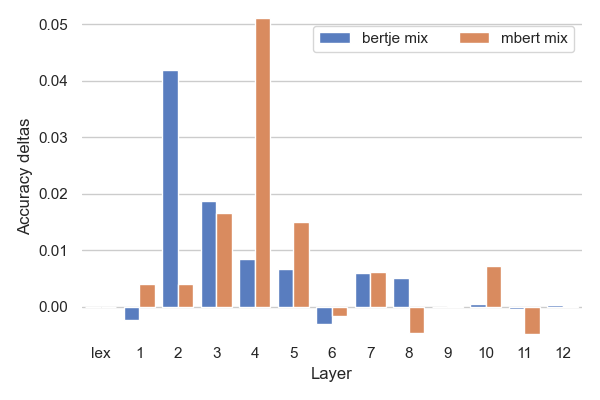}
    \caption{SoNaR Coref}
  \end{subfigure}
  \caption{Accuracy deltas for cumulative introduction of layers with scalar mixing probes. Positive values indicate that these layers contain new task-specific information. Some negative values in later layers  suggest overfitting.}
  \label{fig:mix:acc-deltas}
\end{figure*}

\paragraph{BERTje makes more use of lexical embeddings.}
The curves in Figure~\ref{fig:weights} show that the probes for BERTje give higher weights to the first layer than the mBERT probes. This suggests that the pre-trained context-independent lexical embeddings of BERTje are more informative for these tasks than those of mBERT. 
This makes sense because mBERT word pieces are shared between languages, so there is more word piece level lexical ambiguity in mBERT than BERTje.

The exception to this pattern is the SoNaR coreference task, where the difference between mBERT and BERTje is small. Establishing whether two spans of text corefer requires more context-dependent information in addition to lexical embeddings, whereas the other tasks contain examples where context is not always required.
BERTje does not rely on the lexical layer more strongly than on subsequent layers for this task.

\paragraph{Weights decrease at final layers.}
If the transformer layers continually add information, the final layer would contain most information.
However, information actually decreases after peaking in layers 5 to 9.
The reason may be that the actual output of the model should be roughly the same as the original input.
Therefore generalisations are discarded in favour of representations that map back to actual word pieces.
Generalisations may lead to information loss if they do not correspond to our target tasks, because original information may become less accessible after generalisation.
The first and last lexical layers contain most token identity information. If the probes did not benefit from learned language model representations, we would observe that these layers are the most important to solve the tasks.
However, the weight peaks that we see in between the lexical layers suggest that the language models contain generalisations that are informative for the given tasks.

\paragraph{mBERT peaks earlier than BERTje.}
The weight peak for the mBERT probes is always in an earlier layer than the peaks of equivalent BERTje probes.
These peaks do not correspond with center measures in BERT probing scalar mixing weights of \citet{tenney2019bert}, since single center measures only correspond with peaks if the distribution is roughly normal.

This might suggest differing priorities during pre-training.
Generally, BERTje's weights start to decrease somewhere in the second half of the layers whereas mBERT's peaks are closer to the center.
This suggests that BERTje uses more layers to generalise than to instantiate back to tokens.
The large vocabulary and variety of languages in mBERT may require mBERT to start instantiating earlier with an equal amount of generalisation and instantiation as a result.

\paragraph{POS and DEP results are consistent across datasets.}
The UDLassy and UDAlpino datasets contain equivalent annotations, but the data originates from different text genres.
Their POS curves in Figure \ref{fig:weights:udlassy-pos} and \ref{fig:weights:udalpino-pos} and their DEP curves in Figure \ref{fig:weights:udlassy-dep} and \ref{fig:weights:udalpino-dep} are however mostly the same.
This indicates that the probes are sensitive to the task and the input embeddings, but not overly sensitive to the specific data that the probes are trained on.

\subsection{Prediction scores}\label{sec:predictions}

Figure \ref{fig:mix:acc-deltas} shows deltas of accuracy scores compared to the preceding layer based on test predictions.
The minimum absolute accuracy scores for each task range from 0.630 (SoNaR Coref) to 0.979 (CoNLL-2002 NER) and the maximum accuracy scores per task range from 0.729 (SoNaR Coref) to 0.991 (CoNLL-2002 NER).\footnote{Accuracy deltas for single layer probes are in Appendix~\ref{appendix:accuracies}.}

Intuitively, positive deltas in the mixing results in Figure~\ref{fig:mix:acc-deltas} indicate that the introduced layer contains new information that was not present in any preceding layers, whereas zero-deltas indicate that the new layer is completely uninformative.
Ideally, the accuracy deltas would never be negative since the probe of layer $N$ has access to information from all layers up to $N$.
Negative deltas with cumulative introduction of layers to the probes suggest that the probes sometimes overfit to training data.
Otherwise, these deltas should always be zero or higher.
Scalar mixing weights of layers that correspond with these uninformative negative delta layers should be lower in order to reduce their effect on the predictions.
Figure~\ref{fig:weights} shows that negative accuracy deltas mainly correspond with negative weight slopes.
Therefore, the effects in Figure~\ref{fig:weights} may be stronger in optimally performing probes.

The general pattern in the scalar mixing accuracy deltas in Figure~\ref{fig:mix:acc-deltas} is that  deltas are positive in earlier layers and improvement stops for the last layers.
This fits with the decreasing weights for the last layers in the full scalar mixing model  (Figure~\ref{fig:weights}).

One important difference between the layer mixing probes and the single layer probes is that single layer probes sometimes show negative accuracy deltas while the corresponding accuracy delta is positive for the mixing probe.
Positive mixing probe deltas suggest that new information is introduced or made more accessible, whereas the negative single layer deltas suggest that some information is lost or has been made less accessible by the language model.
Intuitively, this indicates that some information is sacrificed in order to make place for new information in the embedding.
If that is the case, the actual probe prediction mistakes may change between layers even if overall accuracy scores stay the same.

Analysis of scalar mixing weights or accuracy on the whole test data only gives an indication of the sum of information for a task.
However, a more fine-grained error analysis is required to give any indication about what information is retrievable in which layer and what information becomes harder to identify.

\section{In-depth analysis for POS tagging}\label{sec:pos}
\begin{figure}[!b]
\centering
\includegraphics[width=1\linewidth]{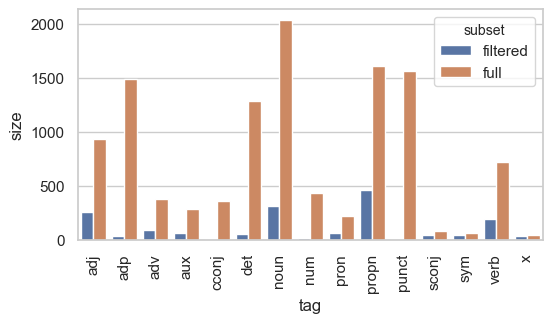}
  \caption{Distributions of POS tags in the full test set as well as the filtered test set. The filtered distribution is not equivalent to the original distribution because some common tags are relatively easy.}
  \label{fig:tag-dist}
\end{figure}

\begin{figure*}[h!]
\centering
\begin{subfigure}[b]{0.45\textwidth}%
\includegraphics[width=\textwidth]{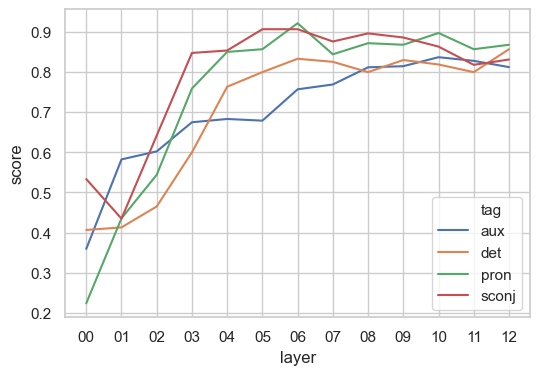}
\caption{BERTje closed class POS tags}
\end{subfigure}
\begin{subfigure}[b]{0.45\textwidth}
\includegraphics[width=\textwidth]{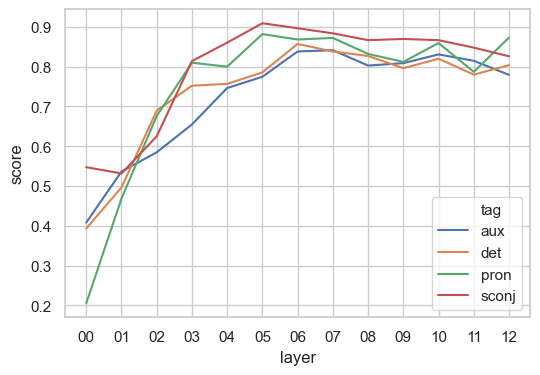}
\caption{mBERT closed class POS tags}
\end{subfigure}
\caption{F1 scores per closed class POS tag per layer for BERTje and mBERT. Closed class performance stabilises around the sixth layers and does not significantly decrease.}
\label{fig:tags:agg:closed}
\end{figure*}

\begin{figure*}[h!]
\centering
\begin{subfigure}[b]{0.45\textwidth}%
\includegraphics[width=\textwidth]{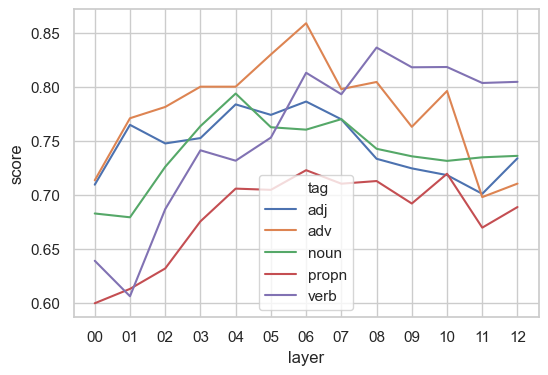}
\caption{BERTje open class POS tags}
\end{subfigure}
\begin{subfigure}[b]{0.45\textwidth}%
\includegraphics[width=\textwidth]{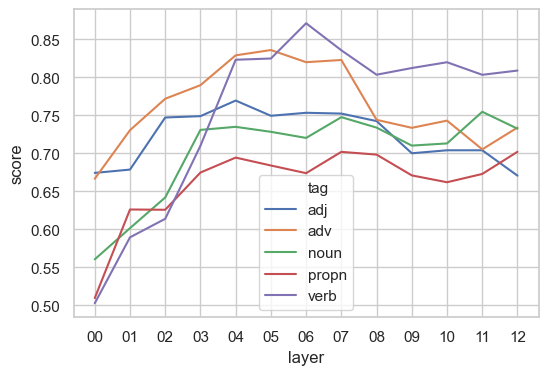}
\caption{mBERT open class POS tags}
\end{subfigure}
\caption{F1 scores per open class POS tag per layer for BERTje and mBERT. Except for verbs, performances decrease in later layers. This indicates that these tag representations become hard to distinguish in later layers.}
\label{fig:tags:agg:open}
\end{figure*}

Layer-wise task performance and scalar mixing weights give information about overall information density for a task.

For POS tagging, maximum performance and largest scalar mixing weights are assigned to layers 5 to 9 for the  pre-trained models, but this does not tell the whole story.
Indeed, probes can make different types of errors for different layers and models, because the models may clarify or lose information between layers.
Moreover, different examples and labels within a task may rely on information from different layers.

We want to give a more thorough view of what BERTje and mBERT learn and whether information becomes unidentifiable between layers as well as whether BERTje and mBERT make the same mistakes.
Therefore, we evaluate the errors of the UDLassy POS predictions with single layer probes.

We do this analysis on POS predictions because this task stays closest to the lexical level of embedding that the models are pre-trained for, but also rely on context and generalisation for optimal performance.
We focus on UDLassy data rather than UDAlpino because the differences between the accuracy deltas of scalar mixing models and single layer models appears higher for UDLassy. This would suggest a larger shift in mistakes.

The following analysis is done on the predictions of the 13 single layer BERTje probes and the 13 single layer mBERT probes.
POS tagging is not difficult for all tokens, so for 85\% of the test data all 26 probes predict the correct tag.
In order to focus on errors, we perform all analyses using the subset of the tokens that have an incorrect prediction by at least one of the probes. This amounts to 1,720 tokens.
The original test data distribution as well as the filtered distribution are shown in Figure~\ref{fig:tag-dist}.

Note that the filtered data distribution does not correspond to the original distribution since some tags are easier to recognise than others.
For instance, proper nouns are over-represented in our analysis set whereas adpositions and punctuation are underrepresented.
This is not a problem since we are explicitly interested in the mistakes and difficult cases and not in overall performance.

\subsection{Accuracies per POS tag}

Figures \ref{fig:tags:agg:closed}~and~\ref{fig:tags:agg:open} show the F1 scores per POS tag per layer for the single layer probe predictions.

POS tags are grouped in aggregates based on whether they are considered to be closed categories (Figure~\ref{fig:tags:agg:closed}) or open categories (Figure~\ref{fig:tags:agg:open}) according to the Universal Dependencies guidelines.
There are six POS tags with relatively low average performance, which also have random fluctuations in per layer performance.
Therefore, \textit{adp}, \textit{cconj}, \textit{punct}, \textit{num}, \textit{sym} and \textit{x} are left out of Figures \ref{fig:tags:agg:closed}~and~\ref{fig:tags:agg:open}.

Figure \ref{fig:tags:agg:closed} shows that closed class POS tags seem to be learned by the pre-trained models and not lost in later layers.
On average, their scores increase for the first six layers, indicating that the probe uses learned information to identify these tags.
After reaching top performance, the probe performance does not really decrease, rather it plateaus.
Only the subordinating conjunction class seems to show some decline.
There is remarkably little difference between BERTje and mBERT for these classes.

Figure \ref{fig:tags:agg:open} shows the tag F1 scores for open class POS tags.
Contrary to the closed classes, the mean scores on open classes do seem to decline in later layers.
Within the closed classes there are three different patterns.
Nouns and proper nouns are learned quickly and stay relatively stable.
This is especially true for  mBERT.
For BERTje, the scores for (proper) nouns seem to decline somewhat after reaching a peak.
Verbs keep improving for more layers than (proper) nouns.
Apparently, recognition of verbs is something that is resolved later in the pre-trained models.
Finally, adjectives and adverbs show an actual decline in performance, since these two tags become hard to distinguish from each other, or possibly other tags, in later layers.

\subsection{Confusion between tags}
The previous figures give an indication about which POS tags are learned by pre-trained models based on context and which tags become unidentifiable, but they do not give an indication about changes in tag confusion.
Figure \ref{fig:tags:agg:open} shows that overall single layer performance of open class words peaks in layer~6 for BERTje and layer~6 is also included in the peak layers for mBERT.

\begin{figure*}
\centering
\begin{subfigure}[b]{0.48\textwidth}
\includegraphics[width=\textwidth]{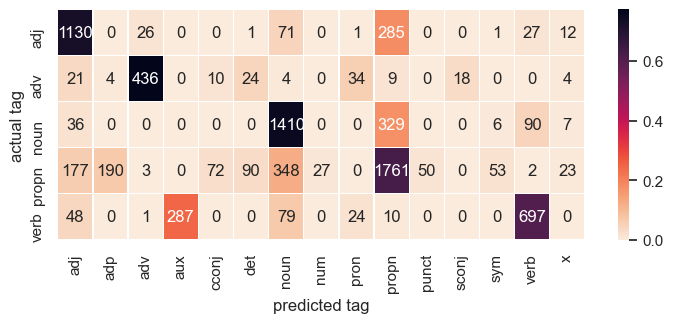}
\caption{BERTje layers 0 up to 5}
\end{subfigure}
\begin{subfigure}[b]{0.48\textwidth}
\includegraphics[width=\textwidth]{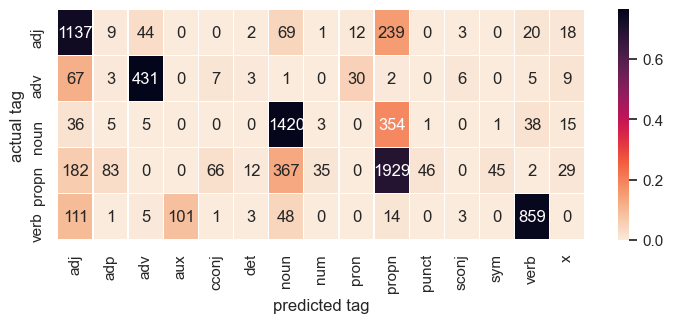}
\caption{BERTje layers 7 up to 12}
\end{subfigure}
\begin{subfigure}[b]{0.48\textwidth}
\includegraphics[width=\textwidth]{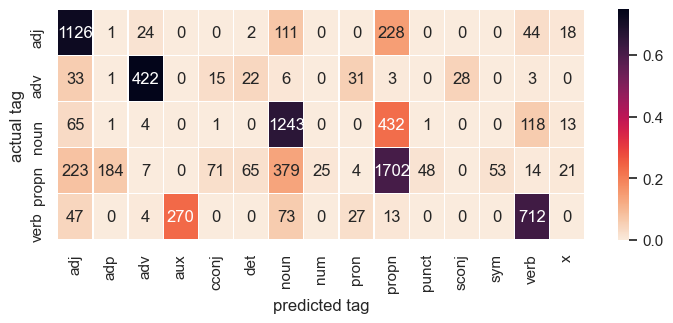}
\caption{mBERT layers 0 up to 5}
\end{subfigure}
\begin{subfigure}[b]{0.48\textwidth}
\includegraphics[width=\textwidth]{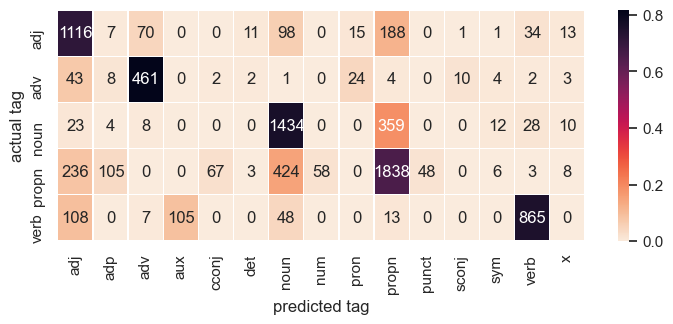}
\caption{mBERT layers 7 up to 12}
\end{subfigure}

\caption{Total confusions of open class POS tags before and after the middle. Confusions are very similar between BERTje and mBERT, but some confusions change between first and last layers.}
\label{fig:cm:summed:open}
\end{figure*}

To illustrate whether biases and confusions change after this peak, we compare the summed confusion matrices from the six layers before and the six layers after layer~6. These confusion matrices (Figure~\ref{fig:cm:summed:open}) show that there are many similarities between BERTje and mBERT with respect to the confusions that are learned or lost.

Decrease in error counts between the first half and the second half of the models suggests that differentiation between tags is learned, whereas increase in errors suggests information loss.
For instance verbs and adverbs are more often misclassified as determiners in the first than in the second half. 
Similarly, proper nouns are confused a lot more often with auxiliary verbs or pronouns in the first half than in the second half.

Those differences suggest that discrimination between these tags is learned by both models.
However, nouns and proper nouns are confused with adjectives a lot more often in the second than in the first half.

\subsection{Example errors}

BERTje and mBERT do not always make the same mistakes, nor are the same mistakes made in each layer.
For many tokens, the probes make incorrect predictions for the first layer(s), but start making correct predictions in later layers, which indicates that learned information is used.
Often, these error patterns are similar between BERTje and mBERT.
The following are examples of differences:
\setlength{\Extopsep}{5pt}

\smallskip

\ex.\label{ex1} Max Rood --- minister van Binnenlandse Zaken , kabinet - Van \textbf{Agt} III \\
    {[}Max Rood --- minister of Internal Affairs , cabinet - Van \textbf{Agt} III{]}

\ex.\label{ex2} \textbf{Federale} Regering \\
    {[}\textbf{Federal} Government{]}

\ex.\label{ex3} Het ontplooiingsliberalisme stelde de vrije \textbf{maar} verantwoordelijke mens centraal. \\
	{[}The self-development liberalism put the free \textbf{but} responsible man central.{]}

\ex.\label{ex4} \textbf{Reeds} in het begin van de 20ste eeuw \dots\\
    {[}\textbf{Already} in the beginning of the 20th century{]}

\ex.\label{ex5} \dots het \textbf{Duitstalig} taalgebied \dots\\
    {[}\dots the \textbf{German} language-area \dots{]}

\ex.\label{ex6} \dots de Keltische \textbf{stammen} in het gebied \dots \\
    {[}\dots the Celtic \textbf{tribes} in the area \dots{]}

\smallskip

\noindent In \ref{ex1}, mBERT initially tags the proper noun ``Agt" as verb. In \ref{ex2} BERTje initially tags the adjective ``Federale" as proper noun.
Both classifications are incorrect guesses, but with additional context both pre-trained models correctly identify this proper noun in later layers.
A different pattern of errors is that the probes make correct predictions based on the first or last layer, but some mistakes for layers in between.
In \ref{ex3} the conjunction ``maar" (but) receives the tag adv in several layers instead of the correct tag ``cconj".
BERTje makes this mistake in layer 4, 5, and 10; mBERT makes it in layers 3 to 7.
It happens relatively often that all BERTje probes assign correct labels, but mBERT goes from incorrect to correct. These mistakes are typically resolved in the first layer of mBERT, suggesting such errors are easily resolvable with a little bit of context; see \ref{ex4} for an example.

There are also a lot of examples where mBERT probes are always correct, but BERTje probes make a mistake somewhere in the middle.
It may be the case that these examples are resolvable with and without context but that the internal representations of BERTje get generalised based on non-POS properties. In \ref{ex5} the adjective ``Duitstalig" gets confused with proper noun in layers 4, 5, 7, 8 and 9, but in the layers before and after BERTje probes get it correct.
Semantically it is reasonable to think that ``Duitstalig" has proper noun-like properties.
Finally, \ref{ex6} is an example where BERTje is always correct but mBERT makes a mistake in the middle somewhere.
The word ``stammen" should be a noun but mBERT sometimes thinks it is a verb.

\section{Conclusion}

Our results show that BERTje and mBERT exhibit a pipeline-like behaviour along tasks similar to what has previously been shown for English.

\citet{tenney2019bert} observed that the pipeline order is roughly first POS tagging, then named entity recognition, then dependency parsing and coreference resolution. 
Our results suggest that BERTje encodes these pipeline tasks in a similar order.
Scalar mixing weights show that there is not a single layer that contains all important information because the weight curves show peaks and valleys.
This suggests that useful task information is distributed between layers.
Generally, the most informative layers are located early in the second half of the pre-trained models.
As an additional note, because we ran the model on different datasets for the same task, we can assess stability across datasets.
We observe that POS tagging and dependency parsing results are consistent, suggesting that the probes are sensitive to the task and the embeddings, but not overly sensitive to the specific data that they are trained on.

The main task differences between the monolingual BERTje model and the multilingual mBERT model are that BERTje probes make more use of the lexical embedding layer than the mBERT probes and the most important layers of BERTje are mostly later layers than those of mBERT.

Semantically rich POS tags like nouns and adjectives become harder to identify in later layers (Figure~\ref{fig:tags:agg:open}) and confusions mainly happen between semantically rich open categories (Figure~\ref{fig:cm:summed:open}). This suggests that semantic content is more important than POS discriminating features for final token predictions.
So even if the POS abstraction is not readily present in the lexical layer nor in the final token prediction layer, POS tag information is still found in middle layer generalisations.
POS tagging is a part of what the pre-trained models learn, but different tag abstractions are present in different layers.
Therefore, feature-based use of these models should not use the output of a \textit{single best} layer.
It would be better to combine the outputs of multiple or all layers in order to retrieve all learned information that is relevant for a downstream task.
However, actual fine-tuning of pre-trained language models should still be a preferred approach.

In sum, our results show that pipeline-like behaviour is present in both a monolingual pre-trained BERT-based model as well as a multilingual model even though task-specific information is distributed between layers.
We observed this for POS tagging, but it is still unclear how information within tasks is distributed in these models for other tasks. 
Moreover, it would be interesting to investigate 
alternative probing strategies in order to better disentangle what pertains to the model itself from what is specific to a given probing strategy. 
Lastly, it is an open question how well linguistic properties are embedded within large pre-trained language models for non Indo-European languages.

\bibliography{main}
\bibliographystyle{acl_natbib}

\appendix

\section{Data}
\label{appendix:data}
This is a more detailed description of the data and data preparation that the probing classifiers are trained and tested on.

For token level classification tasks like POS tagging, the input span is the range of word pieces that form a single token.
For other tasks that use multi-word expressions, like named entity recognition, the spans can be longer than single tokens.
Dependency parsing and coreference resolution are not flat token classification tasks but edge prediction tasks.
Therefore the probing model can also predict edge labels if two spans are given.
The task specific input and output representations are described below.
Table~\ref{tab:data:train} shows the sizes of our training datasets and Table~\ref{tab:data:test} shows the data sizes of our test data. Validation data sizes are nearly the same as test data.

\begin{table*}[t]
	\begin{center}
		\begin{tabular}{l | r r | r r}
			\toprule
			 task & \# sents & \# tokens & \# examples & \# labels \\
			\midrule
			UDLassy POS     &  5,787 &  75,165 &  75,165 & 16 \\
			UDLassy DEP     &  5,787 &  75,165 &  69,293 & 34 \\
			UDAlpino POS    & 12,264 & 185,999 & 185,999 & 16 \\
			UDAlpino DEP    & 12,264 & 185,999 & 173,619 & 34 \\
			CoNLL-2002 NER  & 15,806 & 202,644 & 114,288 &  5 \\
			SoNaR Coref NER & 46,969 & 773,968 & 139,005 &  2 \\
			\bottomrule
		\end{tabular}
	\end{center}
	\caption{\label{tab:data:train} Description of our training data.}
\end{table*}

\begin{table*}[t]
	\begin{center}
		\begin{tabular}{l | r r | r r}
			\toprule
			 task & \# sents & \# tokens & \# examples & \# labels \\
			\midrule
			UDLassy POS     &   875 & 11,581 & 11,581 & 16 \\
			UDLassy DEP     &   875 & 11,581 & 10,681 & 34 \\
			UDAlpino POS    &   596 & 11,053 & 11,053 & 16 \\
			UDAlpino DEP    &   596 & 11,053 & 10,450 & 34 \\
			CoNLL-2002 NER  & 5,195 & 68,875 & 38,488 &  5 \\
			SoNaR Coref NER & 5,094 & 96,705 & 17,720 &  2 \\
			\bottomrule
		\end{tabular}
	\end{center}
	\caption{\label{tab:data:test} Description of our test data. All validation data is in the same order of magnitude as test data.}
\end{table*}

\paragraph{Part-of-speech (POS) tagging}
For POS tagging, two datasets from Universal Dependencies (UD) v2.5 \citep{ud25} are used.
These two datasets are the LassySmall (UDv2.5~LassySmall~POS) and the Alpino (UD~Alpino~POS) datasets, both of which consist of documents from the Lassy Small corpus \citep{lassy}.
The UD-LassySmall data consists of Wikipedia articles whereas the UD-Alpino data originates from news articles.
Universal POS tags are used with 16 coarse lexical categories\footnote{\url{https://universaldependencies.org/u/pos/}}.
Both datasets have predefined train, validation and test splits.

\paragraph{Dependency (DEP) parsing}
For dependency parsing, the same same sources with the same splits are used as for POS tagging: UD-LassySmall and UD-Alpino from UD-v2.5.
For uniformity across tasks, the probing classifiers are not trained for attachment but for edge labeling.
For each edge in a sentence, the head token is used as one span and the full child sequence is used as the other span.
The child span is not a single token since a child forms a semantic unit together with its sub-children.
For instance, a child span can be "A small child" with a head token "plays" where "plays" is the actual head of "child" in the dependency tree.
The semantics of a dependency relationship may be distributed among the tokens within the child tree.
The probing classifier is trained to predict which of the 37 UD syntactic relations is the correct one between the head and child span. Predefined splits are used for training, validating, and testing. 

\paragraph{Named Entity Recognition (NER)}
For Named Entity Recognition, we use the Dutch portion of the CoNLL-2002 NER dataset \citep{tjong2002conll}, which contains BIO-encoded named entity annotations for newspaper articles with four classes: persons, organisations, locations and miscellaneous.
Spans for full entities are used as inputs for the probing classifier with the entity class as target label.
The non-entity tokens are used as negative samples (\textit{O} label) with random span lengths of one to three tokens. The existing train, validation (test1) and test (test2) splits are used.

\paragraph{Coreference (Coref) resolution}
For coreference resolution, the coreference annotations from the SoNaR-1 corpus \citep{sonar1} are used.
There are no pre-defined splits for training and testing, so a random set of 10\% of the documents is used for validation and 10\% for testing.
The splitting is done at document level, so all sentences from the same document are present in the same split.
The coreference task is framed as a binary classification task where two spans of tokens are either coreferential or they are not.
Because referents are often mentioned in multiple sentences, embeddings are extracted from the pre-trained models with concatenated sentences, until the token limit of 512 tokens is reached.
Half of the examples are coreferential strings and half are random referents that do not corefer.
Positive examples are sampled from all possible coreferring spans, whereas negative samples can be any non-coreferring expressions.
The data contains annotations for spans of potentially referring expressions including singletons, so spans in negative examples are not limited to expressions that are coreferential with another span.

\section{Probe hyper-parameters}
\label{appendix:hyperparams}
The probing classifiers use the following hyper-parameters:

\begin{itemize}
    \item Input size: 768 (embedding size of the pre-trained models)
    \item Hidden layer size: 256
    \item Number of bidirectional LSTM layers: 2 (for span representations)
    \item Dropout:
    \begin{itemize*}
        \item Input layer: 0.2
        \item Recurrent layers: 0.3
        \item Other layers: 0.2
    \end{itemize*}
\end{itemize}

This model is trained with the Adam optimisation algorithm with a learning rate of 0.0001 and weight decay of 0.01.
Training is done in mini-batches of 32 examples with evaluation on validation data after every 1000 batches.
Training stops when validation loss has not decreased for 20 steps.

\begin{figure*}[hb]
  \begin{subfigure}[b]{0.5\textwidth}
    \includegraphics[width=\textwidth]{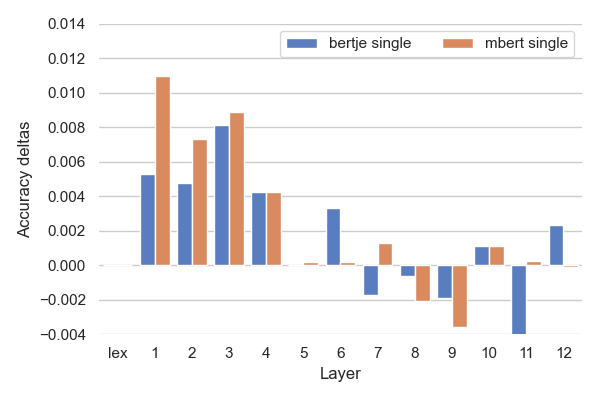}
    \caption{UDLassy POS}
  \end{subfigure}
  \begin{subfigure}[b]{0.5\textwidth}
    \includegraphics[width=\textwidth]{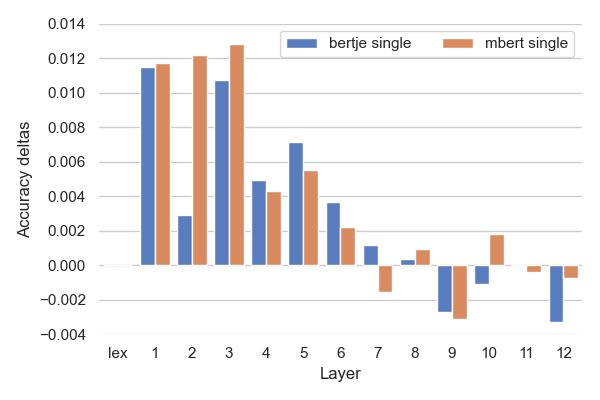}
    \caption{UDAlpino POS}
  \end{subfigure}
  \begin{subfigure}[b]{0.5\textwidth}
    \includegraphics[width=\textwidth]{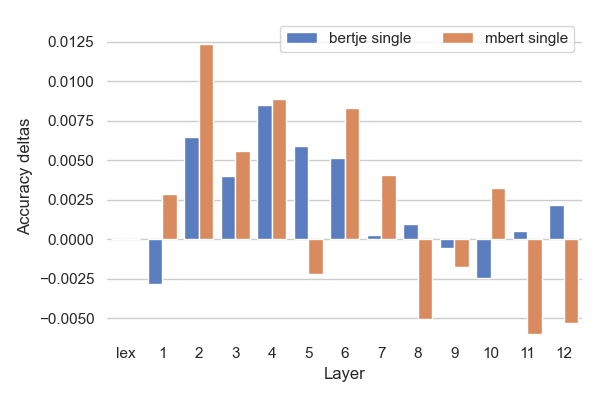}
    \caption{UDLassy DEP}
  \end{subfigure}
  \begin{subfigure}[b]{0.5\textwidth}
    \includegraphics[width=\textwidth]{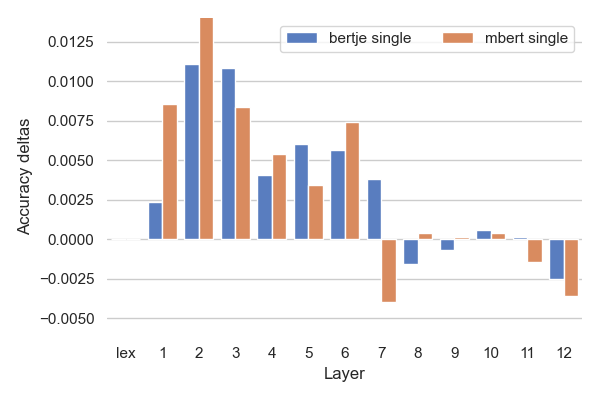}
    \caption{UDAlpino DEP}
  \end{subfigure}
  \begin{subfigure}[b]{0.5\textwidth}
    \includegraphics[width=\textwidth]{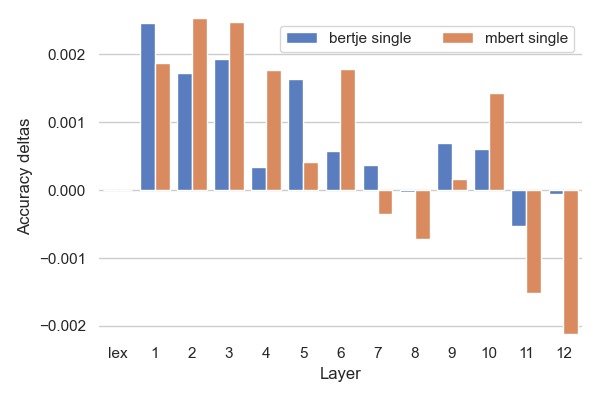}
    \caption{CoNLL-2002 NER}
  \end{subfigure}
  \begin{subfigure}[b]{0.5\textwidth}
    \includegraphics[width=\textwidth]{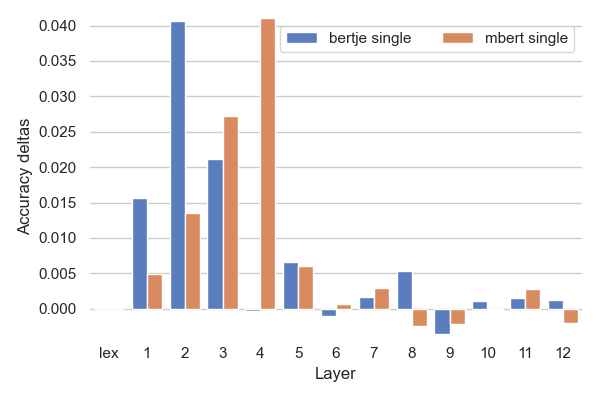}
    \caption{SoNaR Coref}
  \end{subfigure}
  \caption{Accuracy deltas for single layer probes.
  The general pattern is that the deltas are positive in the earlier layers and improvement stops for the last layers.
  }
  \label{fig:single:acc-deltas}
\end{figure*}

\section{Probe accuracies}
\label{appendix:accuracies}
The paper includes accuracy deltas for scalar mixing probes for each task.
Figure~\ref{fig:single:acc-deltas} shows the equivalent accuracy deltas for single layer probes.

\end{document}